\documentclass[sigconf]{acmart}

\usepackage{enumitem}
\usepackage{ subfigure }
\newtheorem{prop}{Proposition}

\begin{document}
\vspace{-1em}
\copyrightyear{2020}
\acmYear{2020}
\setcopyright{acmcopyright}\acmConference[CIKM '20]{Proceedings of the 29th ACM International Conference on Information and Knowledge Management}{October 19--23, 2020}{Virtual Event, Ireland}
\acmBooktitle{Proceedings of the 29th ACM International Conference on Information and Knowledge Management (CIKM '20), October 19--23, 2020, Virtual Event, Ireland}
\acmPrice{15.00}
\acmDOI{10.1145/3340531.3412139}
\acmISBN{978-1-4503-6859-9/20/10}
% Authors, replace the red X's with your assigned DOI string during the rightsreview eform process.

\settopmatter{printacmref=true}

\fancyhead{}
\title{Label-Aware Graph Convolutional Networks}
\vspace{-1em}
	\author{Hao Chen$^{1\dagger}$, Yue Xu$^{2\dagger}$, Feiran Huang$^{3}$, Zengde Deng$^{4}$, Wenbing Huang$^{5}$,}
	\author{Senzhang Wang$^{6}$, Peng He$^{7}$ and Zhoujun Li$^{1\ddagger}$}
	\affiliation{\institution{$^{1}$Beihang University, $^{2}$Alibaba Group, $^{3}$Jinan University, $^{4}$CUHK, $^{5}$Tsinghua University,}\institution{$^{6}$Nanjing University of Aeronautics and Astronautics, $^{7}$Wechat, Tencent Inc.}
	%\city{City}\state{State}
	}\email{sundaychenhao@gmail.com, guyue.yuexu@alibaba-inc.com, huangfr@jnu.edu.cn, dengzengde@gmail.com,}\email{hwenbing@126.com, szwang@nuaa.edu.cn, paulhe@tencent.com, lizj@buaa.edu.cn}

% 	\author{1 \\E1\vspace{-2cm}
% \thanks{xxx}\\
% \and 2\\E2\vspace{-2cm}\\
% \and A3\\E3\vspace{-2cm}\\
% \and A4\\E4\vspace{-2cm}\\}

% 	% author
% 	\author{Yue Xu}
% 	\affiliation{\institution{Institute}
% 	}\email{author@example.com}
	
% 	% author
% 	\author{Third Author}
% 	\affiliation{\institution{Institute}
% 	}\email{author@example.com}
	
% 	% author
% 	\author{Forth Author}
% 	\affiliation{\institution{Institute}
% 	}\email{author@example.com}
	
% 	% author
% 	\author{Fifth Author}
% 	\affiliation{\institution{Institute}
% 	}\email{author@example.com}
	
% 	% author
% 	\author{Sixth Author}
% 	\affiliation{\institution{Institute}
% 	}\email{author@example.com}
% 		% author
% 	\author{Sixth Author}
% 	\affiliation{\institution{Institute}
% 	}\email{author@example.com}
% 		% author
% 	\author{Sixth Author}
% 	\affiliation{\institution{Institute}
% 	}\email{author@example.com}

	\begin{abstract}
		Recent advances in Graph Convolutional Networks (GCNs) have led to state-of-the-art performance on various graph-related tasks. 
		However, most existing GCN models do not explicitly identify whether all the aggregated neighbors are valuable to the learning tasks, which may harm the learning performance.
		In this paper, we consider the problem of node classification and propose the Label-Aware Graph Convolutional Network (LAGCN) framework which can directly identify valuable neighbors to enhance the performance of existing GCN models.  Our contribution is three-fold.
		First, we propose a label-aware edge classifier that can filter distracting neighbors and add valuable neighbors for each node to refine the original graph into a label-aware~(LA) graph. Existing GCN models can directly learn from the LA graph to improve the performance without changing their model architectures.
		Second, we introduce the concept of positive ratio to evaluate the density of valuable neighbors in the LA graph. Theoretical analysis reveals that using the edge classifier to increase the positive ratio can improve the learning performance of existing GCN models.
		Third, we conduct extensive node classification experiments on benchmark datasets. The results verify that LAGCN can improve the performance of existing GCN models considerably, in terms of node classification.\let\thefootnote\relax\footnotetext{\noindent$\dagger$ Denotes equal contributions.\\
		$\ddagger$ Corresponding author.\\Work done when first author interns at Wechat Tencent Inc..}
	\end{abstract}
% 	\vspace{-1em}
% 	\vspace{-1em}
	

	\maketitle

% 	\vspace{-1em}
	\section{Introduction}
	\vspace{-0.2em}
	Node classification, which aims at classifying nodes in a graph, is one of the most common applications in graph-related literatures~\cite{bhagat2011node,perozzi2014deepwalk,grover2016node2vec,wang2014mmrate}, e.g., classifying the documents in a citation network. 
	Recently, the Graph Convolutional networks (GCNs) which extend the Convolutional Neural Networks~(CNNs) to the non-Euclidean graph domain are gaining increasing research interests.
	They have been successfully applied to node classification and achieved the state-of-the-art performances~\cite{kipf2016gcn,velivckovic2017gat,zhang2018gaan,huang2018asgcn,chen2018fastgcn,zeng2019graphsaint,rong2019dropedge}.
	These successes also encourage the burst of many variants, including GraphSAGE~\cite{hamilton2017graphsage},  simplified GCN (SGC)~\cite{wu2019sgc}, ASGCN~\cite{huang2018asgcn}, FastGCN~\cite{chen2018fastgcn}, and GAT~\cite{velivckovic2017gat}, etc.
	
	Generally, in most GCN models, the state of each node is iteratively updated by its own state and the states aggregated from its neighbors~\cite{li2018laplacian,wu2019sgc}. 
	The final output (e.g., the predicted label of the node class) is produced based on the aggregated node states.
	As such, the performance of GCN models is largely effected by the quality of the aggregated neighbors.
	However, existing GCN models do not explicitly identify whether the aggregated neighbors are valuable to the node classification or investigate how to enhance the accuracy of GCN models by refining the graph structure. In this case, the classification result may be misled by the information aggregated from distracting neighbors.
	
	A number of recent works generalize the graph convolution with the attention mechanism~\cite{velivckovic2017gat,zhang2018gaan}. For example, GAT~\cite{velivckovic2017gat} dynamically assigns an aggregation weight to each neighbor, which, to some extent, can control the aggregation from distracting neighbors. However, the attention layers are not trained with an explicit criterion to identify whether one neighbor is informative or distracting; besides, the attention layers cannot help one node to gain information from other non-connected but valuable nodes. Consequently, the attention mechanism may have little effect when the target node has very few informative neighbors. In addition, it is also hard for other GCN models with different architectures to reuse the trained attention weights directly.
	
	In this paper, we propose the Label-Aware GCN (LAGCN) framework which can refine the graph structure before the training of GCN, such that existing GCN models can directly benefit from LAGCN without changing their model architectures. The main contribution is three-fold.
% 	\vspace{-0.7em}
	\begin{itemize}[leftmargin=*]
		\item We point out and verify that in node classification tasks, the same-label neighbors are valuable~(positive) neighbors while the different-label neighbors are distracting~(negative) neighbors. 
		Then we propose an edge classifier to refine the original graph structure into a label-aware~(LA) graph structure by 1)~\textit{filtering} the negative neighbors and 2)~\textit{adding} more non-connected but positive nodes as new neighbors. Existing GCN models can be trained directly over the LA graph to improve their performance, without changing their model architectures.
		\item We introduce the concept of positive ratio to evaluate the density of valuable neighbors in the LA graph and prove that the performance of GCN is positively correlated to the positive ratio from a theoretical standpoint. On this basis, we give the minimum requirements of building an effective edge classifier and reveal how the edge classifier influences the classification performance.
		\item We conduct extensive experiments on benchmark datasets and verify that the LAGCN can improve the node classification performances of existing GCN models (including GAT) considerably, especially when the underlying graph has a low positive ratio.
	\end{itemize}
	\vspace{-1em}

	\section{Preliminaries}
	In this paper, we consider the node classification problem on a directed graph $\textbf{G}=(\textbf{V},\textbf{E})$ with $N$ nodes and $E$ edges. 
	The edge between node $ u,v \in \textbf{V}$ is denoted as $(u,v)\in\textbf{E}$. 
	The binary adjacency matrix with self-loop edges is denoted as $\textbf{A}\in\mathbb{R}^{N\times N}$.
	Let $\textbf{X} = \{\textbf{x}_1, \textbf{x}_2, \cdots, \textbf{x}_N\}$ denote the feature of the nodes and $Y=\{y_1, y_2, \cdots, y_N\}$ denote the labels of all the nodes. 
	Let $C$ denotes the total class number, $\mathcal{N}_v$ denotes the set of all $ 1 $-hop neighbors of node $v$ and $n_v=|\mathcal{N}_v|$ denotes the number of $ 1 $-hop neighbors of node $v$. 
	The hidden embedding of node $v$ on the $k$-th GCN layer is denoted as $\textbf{h}_v^{(k)}$.
	The initial node embedding at the first layer is generated with the raw feature $\textbf{X}$.

	% 	In what follows, we briefly review several state-of-the-art GCN models which are also the compared baselines in Sec.~\ref{sec:experiment}.
	Generally, most existing GCNs are constructed by stacking multiple first-order graph convolution layers. Specifically, the embedding of node $v$ at the $(l+1)$-th layer of an GCN can be computed as
	\begin{equation}
	\textbf{h}_{v}^{(l+1)} = \sigma\Big( \textbf{FC}_{\theta_{(l)}} \big(\sum\nolimits_{u\in{\mathcal{N}_v}} w^{(l)}_{u,v} \textbf{h}^{(l)}_u \big)\Big),
	\label{equ:agg}
	\end{equation}
	where $\textbf{FC}_{\theta_{(l)}}$ refers to a fully-connect layer with the parameter $ \theta_{(l)} $; $w^{(l)}_{v,u}$ denotes the aggregation weight of the neighboring nodes $u\in{\mathcal{N}_v}$; $\sigma$ denotes the non-linear activation function. 
	In GAT~\cite{velivckovic2017gat}, the aggregation weights~(i.e., attention weights) are determined based on a multi-head attention mechanism. While in sampling-based GCNs~\cite{hamilton2017graphsage,chen2018fastgcn,huang2018asgcn}, the central node $v$ samples a subset of the neighbors $\mathcal{N}_v'$ for aggregation based on certain pooling methods, e.g., mean pooling.

	Noticeably, the recently proposed Simplified Graph Convolutional Networks~(SGC)~\cite{wu2019sgc} reveals that the nonlinearity between consecutive convoulutional layers can be unnecessary. 
	Hence, SGC removes the non-linear activation functions such that the node embedding at the $(l+1)$-th layer of SGC can be simplified as
	\begin{equation}
	\textbf{h}_{v}^{(l+1)} = \frac{1}{n_v} \sum\nolimits_{u\in{\mathcal{N}_v}} \textbf{h}^{(l)}_u .
	\label{equ:sgc}
	\end{equation}
	After computing the hidden embedding of $K$-th layer, SGC predicts the labels of nodes with a single linear layer and achieves comparable performance as GCN. 
%	In the next section, we use SGC to better demonstrate the motivation behind LAGCN.
% 	\vspace{-1em}
	\section{Lable-Aware GCN Framework}
	In this section, we introduce the concept of positive ratio, discuss how to build an effective edge classifier, and analyze the correlation between the positive ratio and classification performance from a theoretical standpoint.
% 	\vspace{-1em}
% 	\subsection{Positive Ratio}
	\begin{figure}[t]
	\begin{center}
		\centerline{\includegraphics[width=\linewidth]{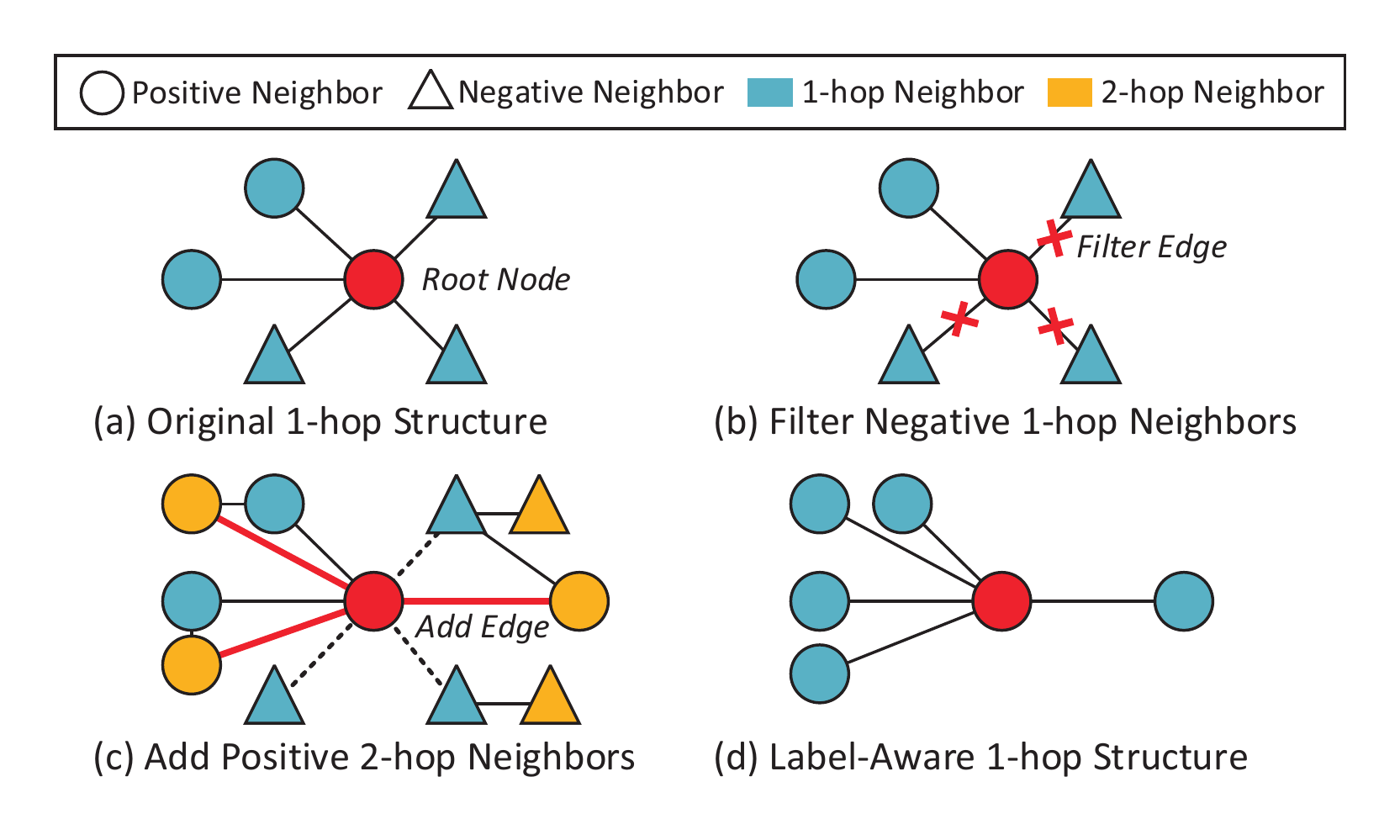}}
		\vskip -2em
		\caption{Illustration of the graph refinement process. 
		}
		\label{fig:framework}
	\end{center}
	\vskip -3em
	\end{figure}
	Given a target node, its neighbors can be classified into two types, i.e., the positive ones and the negative ones.
	\begin{definition}\label{def1} 
		Given a target node $v$, its positive (or negative) neighbor set $\mathcal{N}^+_v$ (or $\mathcal{N}^-_v$) is the set of neighbors whose labels $y_u$ are the same as (or different from) $y_v$.
%		~\footnote{This can also be generalized to multi-label datasets by defining the similarity among nodes, which can be studied in future works.}.
		%The positive ratio is defined as the ratio of positive neighbors.
	\end{definition}
	\noindent As such, the update function of node $v$ in~\eqref{equ:agg} can be rewritten as
	\begin{equation}
	\textbf{h}_{v}^{(l+1)} = \sigma \Big(\textbf{FC}_{\theta_{(l)}}(\sum\nolimits_{u\in{\mathcal{N}^+_v}} w^{(l)}_{v,u} \textbf{h}_{u}^{(l)} + \sum\nolimits_{u\in{\mathcal{N}^-_v}} w^{(l)}_{v,u} \textbf{h}_{u}^{(l)}) \Big).
	\end{equation}
	According to SGC~\cite{wu2019sgc}, the nonlinearity between GCN layers can be unnecessary since the main benefits come from the local averaging. 
	Therefore, we here develop our theoretical analysis without considering the nonlinear functions (including the fully-connected layer and the activation function) in GCNs for analytical purposes. 
	In this case, the update function of node $v$ can be written as
	\begin{equation}
	\textbf{h}_v^{(l)} = \frac{1}{n_v}(\sum\nolimits_{u\in{\mathcal{N}^+_v}}  \textbf{h}_{u}^{(l-1)} + \sum\nolimits_{u\in{\mathcal{N}^-_v}} \textbf{h}_{u}^{(l-1)}).
	\label{equ:sgc_agg}
	\end{equation}
	With the simplified graph convolution, the embedding of different layers locates at the same hidden space. 
	Next, we demonstrate that the probability of generating a correct classification result is correlated to the positive ratio. 
	Without loss of generality, we consider a binary classification problem with the following assumptions\footnote{Note that the assumptions of binary classification and Gaussian distribution are not necessary for real applications, we use them here only for analytical purpose.}.
	\begin{itemize}[leftmargin=*]
		\item Given a linear function $F$ which projects the hidden embedding $\textbf{h}_u^{(l-1)}$ to a real number prediction of node $u$, such as the linear function of SGC. We classify one node $u$ into the positive class if $F(\textbf{h}_u^{(l-1)}) > \tau$, or negative class otherwise, where $\tau$ denotes the classification threshold.
		\item We assume that the prediction of $F(\cdot)$ obeys a Gaussian mixture distribution, i.e.,  $F(\textbf{h}_u^{(l-1)}){ iid \atop \sim} Norm(\mu^+, \sigma^2)$ for $u \in \mathcal{N}^+_i$ and $F(\textbf{h}_u^{(l-1)}){ iid \atop \sim} Norm(\mu^-, \sigma^2)$ for $u \in \mathcal{N}^-_i$. 
		Empirically, we consider $\mu^+$ to be larger than $\tau$, and $\mu^-$ otherwise.
	\end{itemize}
	In this way, $F(\textbf{h}_v^{(l)})$ can be written as 
	\begin{equation}
	F(\textbf{h}_v^{(l)}) = \frac{1}{n_v}
	\Big(\sum\nolimits_{u\in{\mathcal{N}_v^+}} F(\textbf{h}_u^{(l-1)})
	+ \sum\nolimits_{u\in{\mathcal{N}_v^-}} F(\textbf{h}_u^{(l-1)}) \Big).
	\end{equation}
	%{\color{red}Note that $F(\textbf{h}_v^{(l)})$ is a linear combination of several Gaussian random variables and therefore {\color{blue}follows} the Gaussian distribution.}
	Then the expectation of $F(\textbf{h}_v^{(l)})$ can be written as 
	\begin{equation}
	E_{origin}=E(F(\textbf{h}_v^{(l)})) = \dfrac{1}{n_v}(n^+_v \mu^+ + n^-_v \mu^-) = r_v \mu^+ + (1-r_v)\mu^-,
	\end{equation}
	where $r_v = n_v^+/(n_v^+ + n_v^-) = n_v^+/n_v$ denotes the positive ratio of node $v$. 
	% Enlarge $\mathbb{E}(F(\textbf{h}_v^{(l)}))$ can result in a higher probability to classify node $v$ to the positive class.
	Therefore, we can enhance the probability of correctly classifying node $v$ by increasing its positive ratio~$r_v$. Similarly, we can enhance the probability of correctly classifying each node in a graph by increasing the positive ratio of the entire graph, which leads to the following proposition.
	\begin{prop}\label{the:origin}
		The probability of correctly classifying the nodes $v \in \textbf{V}$ in a graph can be increased by improving the positive ratio of the entire graph $R=\sum_{v \in \textbf{V}} n_v^+/(\sum_{v \in \textbf{V}} n_v)$.
	\end{prop}
	\vspace{-1em}
% 	\subsection{Label-Aware Graph Refinement}
	\textbf{Label-Aware Graph Refinement.} We now build an edge classifier to identify positive and negative neighbors for each node in a graph.
	In particular, we refer to the edges between the node with the positive neighbors (or negative neighbors) as a \textit{positive edge} (or \textit{negative edge}).
	Given an edge $(u, v)$, the aim of the edge classifier $\mathcal{E}$ is to return a binary value $\hat{y}_{u,v}$ which specifies whether the edge $(u, v)$ is a positive edge, i.e., $\hat{y}_{u,v}=1$ or a negative edge, i.e., $\hat{y}_{u,v}=0$. 
	The edge classifier $\mathcal{E}$ can be readily trained using the binary adjacency matrix $\textbf{A}$, the node feature matrix $\textbf{X}$, and the labels of the train set $Y_{train}$ in the original graph.
	
	%The edge classifier should be easily computed and satisfy the commutative property which means exchanging the order of the edge will not influence the prediction (i.e., $\mathcal{E}(u,v) = \mathcal{E}(v,u)$).
	In this paper, we build the edge classifier with multi-layer perception (MLP) layers.
	Specifically, given the features of two nodes as $\textbf{e}_u$ and $\textbf{e}_v$, the binary value of their edge is computed as
	\vspace{-0.5em}
	\begin{equation}
	\mathcal{E}(u,v) = \textbf{MLP}\left\lbrace |\hat{\textbf{e}}_u - \hat{\textbf{e}}_v| \ \oplus \ (\hat{\textbf{e}}_u + \hat{\textbf{e}}_v) \ \oplus \ (\hat{\textbf{e}}_u \circ \hat{\textbf{e}}_v)\right\rbrace ,
	\end{equation}
	\vspace{-0.3em}
	where $\oplus$ denotes the concatenation operation, $\textbf{MLP}$ denotes the function of multi-layer perception, and $\hat{\textbf{e}}_u =\textbf{e}_u \textbf{W}_e$ denotes the low-dimensional feature projected from $ \textbf{e}_u $ to accelerate the prediction process with $\textbf{W}_e$ the projecting weights. 
	Note that using MLP layers to build an edge classifier is only one possible solution, other designs could be explored in the future.
%	Given the features $\textbf{e}_u, \textbf{e}_v$ of two nodes $ u $ and $ v $, we use $|\textbf{e}_u - \textbf{e}_v|$,  $\textbf{e}_u + \textbf{e}_v$, and $\textbf{e}_u \circ \textbf{e}_v$ as the input features to train an edge classifier.
%	Moreover, in order to accelerate the prediction process under high-dimensional features, we use a weighted matrix  to project the input features into a space with lower dimensions, i.e., 
%	As such, the edge classifier 
	
	The original graph structure can then be refined into a label-aware graph structure to increase the positive ratio~$R$ based on the edge classifier. 
	Specifically, the graph refinement process contains the following two steps. 
	1) \textbf{Filtering process:} as shown in~Figure \ref{fig:framework}(b), for each node in a graph, we use the trained edge classifier to predict whether its 1-hop neighbors are positive neighbors, then we delete all of its negative neighbors.
	2) \textbf{Adding process}: as shown in~Figure \ref{fig:framework}(c), for each node, we use the trained edge classifier to predict whether its 2-hop neighbors are positive neighbors, then we add edges between this node and its positive 2-hop neighbors, turning them into positive 1-hop neighbors, until the total number of 1-hop neighbors reaches a preset number $n_{max}$.
	%\footnote{Practically, one can sort the positive neighbors according to the confidence level of the prediction results.} 
%	Generally, if the edge classifier is good enough, such graph refinement can 
	
% 	\subsection{Theoretical Analysis}
	\textbf{Theoretical Analysis.} We now analyze why learning from the LA-graph leads to superior learning performance and give the minimum requirements for building an effective edge classifier.
	
	In the \textbf{filtering process}, we delete all the predicted negative edges ($\hat{y}_{u,v} =0$) between the central node $v$ and its 1-hop neighbors. 
	The filtering process preserves two types of neighbors: 
	1) the positive neighbors which are predicted to be positive, i.e., $y_{u,v} = 1$ and $ \hat{y}_{u,v} = 1$; 
	2) the negative neighbors which are predicted to be positive, i.e., $y_{u,v} = 0$ and $\hat{y}_{u,v} = 1$. 
	Suppose that before the filtering process, the number of positive (or negative) neighbors for node $v$ is $n_v^+$ (or $n_v^-$). 
	Then, after filtering the predicted-negative edges, the number of positive (or negative) neighbors for node $v$ turns to be $p\cdot n_v^+$ (or $q\cdot n_v^-$) where $p={\rm P}(\hat{y}_{u,v} =1|y_{u,v} = 1) $ and $q={\rm P}(\hat{y}_{u,v} =1|y_{u,v} =0)$.
	As such, the expectation of  $F(\textbf{h}_v^{(l)})$ can be given as
	\begin{equation}
	E_{filter} = \frac{p\cdot n^+_v \mu^+ + q\cdot n^-_v \mu^-}{p\cdot n_v^+ + q\cdot n_v^-}.
	\end{equation}
	In order to ensure $E_{filter}$ to be greater than $E_{origin}$, the edge classifier should satisfy the following proposition.
	\begin{prop}
		Let $p={\rm P}(\hat{y}_{u,v} =1|y_{u,v} = 1) $ and $q={\rm P}(\hat{y}_{u,v} =1|y_{u,v} =0)$, the filtering process can enhance the performance of the GCN models as long as $p > q$.
		\label{the:filter}
	\end{prop}
	% 	\begin{table*}
	% 		\renewcommand\arraystretch{1.1}
	% 		\small
	% 		\centering
	% 		\caption{Dataset statistics of the origin-Graph and the LA-Graph}
	% 		\vskip -1em
	% 		\setlength{\tabcolsep}{2.5mm}{
	% 			\begin{tabular}{lcccc}
	% 				\toprule
	% 				\textbf{Dataset} & \textbf{Nodes/Edges/Features} & \textbf{SemiTrain/Train/Val/Test} & \textbf{ori-Pos/ori-Neg/ori-Ratio} & \textbf{LA-Pos/LA-Neg/LA-Ratio} \\
	% 				\midrule
	% 				Cora  & 2,708/5,429/1,433 & 140/1,208/500/1,000 & 18.4K/3.2K/85\% & 20.3K/2.2K/90\% \\
	% 				Citeseer & 3,327/4,732/3,703 & 120/1,812/500/1,000 & 6.8K/2.4K/74\% & 16.6K/3.7K/82\% \\
	% 				Pubmed & 19,717/44,338/500 & 60/18,217/500/1,000 & 71.1K/17.5K/80\% & 483.0K/20.4K/96\% \\
	% 				Reddit & 233.0K/11.6M/602 & -/152K/24K/55K & 18.1M/5.1M/78\% & 18.0M/1.0M/95\% \\
	% 				\bottomrule
	% 			\end{tabular}
	% 		}
	% 		\label{tab:dataset}
	% 		\vskip -1em
	% 	\end{table*}
	
	In the \textbf{adding process}, we iteratively add edges between the central node $v$ and its predicted-positive 2-hop neighbors ($\hat{y}_{u,v} =1$), i.e., turning them into 1-hop neighbors, until the total number of 1-hop neighbors reaches a preset number. 
	% In the similar way as the filtering process, there are two types of neighbors that will be added after the adding process: 1) the predicted positive neighbor which is actually a positive neighbor ($\hat{y}_{u,v} =1,y_{u,v} = 1$), 2) the predicted positive neighbor which is actually a negative neighbor ($\hat{y}_{u,v} =1, y_{u,v} = 0$). 
	Suppose that before the adding process, the number of positive (or negative) neighbors of node $v$ is $n_v^+$ (or $n_v^-$) and the number of added 1-hop neighbors is $n_v'$.
	After the adding process, the number of its positive and negative neighbors turns to be $n_v^+ + p_{pre} \cdot n_v'$ and $n_v^- + (1-p_{pre}) \cdot n_v'$, respectively, where $p_{pre}={\rm P}(y_{u,v} = 1|\hat{y}_{u,v} =1)$ and $1-p_{pre}={\rm P}(y_{u,v} = 0|\hat{y}_{u,v} =1)$. 
	The expectation of the prediction $F(\textbf{h}_v^{(l)})$ can be written as
	\begin{equation}
	E_{add} = \frac{ (n^+_v+p_{pre}\cdot n_v') \mu^+ + (n^-_v + (1- p_{pre})\cdot n_v' )\mu^-}{n_v^+ + n_v^- + n_v'}.
	\end{equation}
	In order to ensure $E_{add}$ to be greater than $E_{origin}$, an effective edge classifier should satisfy the following proposition.
	\begin{prop}
		Let $p_{pre}={\rm P}(y_{u,v} = 1|\hat{y}_{u,v} =1)$, the adding process can enhance the performance of GCN as long as $p_{pre} > r_v$. 
		\label{the:add}
	\end{prop}
	\vspace{-0.7em}
	\section{EXPERIMENT} 
	\label{sec:experiment}
% 	\vspace{-0.3em}
% 	\subsection{Experimental Settings}
	\textbf{Datasets.} We use four benchmark datasets, i.e., Cora, Citeseer, Pubmed and Reddit~\cite{hamilton2017graphsage}, where the number of nodes scales as $O(10^3)$, $O(10^3)$, $O(10^4)$, and $O(10^5)$, respectively.
	\\
	\textbf{Settings.} The edge classifier is trained with the nodes and edges in the training set. 
	For Pubmed, we use the raw feature as the input features for edge classifier. For Cora, Citeseer, and Reddit, we use $\textbf{A}^2 \textbf{X}$~\cite{wu2019sgc} as the input features for edge classifier.
	For the adding process, we set the maximum number of 1-hop neighbors as $6$ for both Cora and Citeseer, and as $30$ for Pubmed and Reddit. The maximum numbers are selected to be $20\%$ larger than the sampling number of GraphSAGE~\cite{hamilton2017graphsage}. Code will be released later to ensure reproducibility.
	\\
	\textbf{Baseline models.} We select GCN~\cite{kipf2016gcn}, GAT~\cite{velivckovic2017gat}, and SGC~\cite{wu2019sgc} GraphSAGE~\cite{hamilton2017graphsage} and ASGCN \cite{huang2018asgcn} as the baseline models. 
	For Cora, Citeseer and Pubmed, GCN, GAT, and SGC are trained with semi-supervised setting, i.e., only use a small part of nodes in the training set to optimize their parameters, while GraphSAGE and ASGCN are trained with full-supervised setting. 
	For Reddit, SGC, GraphSAGE, and ASGCN are trained with full-supervised setting.
	
	%The compared supervised methods include . 
	%For all the compared models, we fix the random seeds and use the early stopping strategy (using a window size of $30$ as suggested in~\cite{kipf2016gcn}) to generate the best performances. The present performances are averaged over multiple runs to give a fair comparison.
	
	%To reduce the performance fluctuation, we run these algorithm with 5 different random seed and present the average result.
% 	\vspace{-1em}
% 	\subsection{Experimental Results}    
% 	\vspace{-0.5em}
	\begin{table}[t]
		\small
		\centering
		\caption{Accuracy of the LA-GCNs against the origin-GCNs}
		\vskip -1em
		\resizebox{0.7\columnwidth}{!}{
		\setlength{\tabcolsep}{1mm}{
		\begin{tabular}{lcccc}
			\toprule
			& \textbf{Cora} & \textbf{Citeseer} & \textbf{Pubmed} & \textbf{Reddit} \\
			\midrule
			% 			\textbf{Semi-Supervised Methods} &       &       &  & \\\midrule
			ori-R/LA-R & 85\%/90\% & 74\%/82\% & 80\%/96\% & 78\%/95\% \\
			%& 0.8330 & 0.7330 & \textbf{0.8780} & - \\
			\midrule[0.8pt]
			\midrule[0.8pt]
			origin-GCN & 0.8180 & 0.7090 & 0.7850 & - \\
			LA-GCN & 0.8330 & 0.7330 & \textbf{0.8780} & - \\
			\midrule
			origin-GAT & 0.8300 & 0.7250 & 0.7900 & -\\
			LA-GAT & 0.8350 & \textbf{0.7360} & 0.8690 &-\\
			\midrule
			origin-SGC & 0.8210 & 0.7190 & 0.7890  & 0.9488 \\
			LA-SGC & \textbf{0.8380} & 0.7340 & 0.8770  & 0.9540\\
			\midrule[0.8pt]
			\midrule[0.8pt]
			% 			\textbf{Supervised Methods} &       &       &  \\\midrule
			origin-SAGE & 0.8650 & 0.7850 & 0.8830 & 0.9540 \\
			LA-SAGE & 0.8840 & 0.8000 & 0.9070 & 0.9673\\
			\midrule
			origin-ASGCN & 0.8740 & 0.7960 & 0.9060 & 0.9627\\
			LA-ASGCN & \textbf{0.8880} & \textbf{0.8010} & \textbf{0.9170} & \textbf{0.9758} \\
			\bottomrule
		\end{tabular}}}
		\vskip -0.5em
		\label{tab:overall}
	\end{table}
	
	\begin{table}[t]
		\vskip -0.5em
		\small
		\centering
		\caption{Performance under low positive ratio.}
		\vskip -1em
		\resizebox{0.6\columnwidth}{!}{
		\setlength{\tabcolsep}{0.5mm}{
		\begin{tabular}{lccc}
			\toprule
			& \textbf{Cora} & \textbf{Citeseer} & \textbf{Pubmed} \\
			\midrule
			ori-R(low)/LA-R(low) & 30\%/67\% & 24\%/74\% & 34\%/95\% \\
			\midrule[0.8pt]
			\midrule[0.8pt]
			origin-GCN(low) & 0.3920 & 0.3820 & 0.5810 \\
			LA-GCN(low) & 0.6690 & 0.6650 & 0.8780 \\
			\midrule
			origin-GAT(low) & 0.4430 & 0.3580 & 0.6020 \\
			LA-GAT(low) & 0.6850 & 0.6750 & 0.8850 \\
			\bottomrule
		\end{tabular}}}
		\vskip -2em
		\label{tab:low-homophily-ratio}
	\end{table}
	
	% 	\begin{table}[t]
	% 		\vskip -1.5em
	% 		\small
	% 		\centering
	% 		\caption{Performance of the edge classifier.}
	% 		\vskip -1em
	% 		\begin{tabular}{lcccc}
	% 			\toprule
	% 			& \textbf{Cora} & \textbf{Citeseer} & \textbf{Pubmed} & \textbf{Reddit} \\
	% 			\midrule
	% 			$p$ & 85\%  & 89\%  & 92\%  & 98\% \\
	% 			$q$ & 36\%  & 54\%  & 31\%  & 13\% \\
	% 			$p-q$ & 49\%  & 34\%  & 61\%  & 85\% \\
	% 			$p_{pre}$ & 93\%  & 82\%  & 93\%  & 97\% \\
	% 			Accuracy & 86\%  & 75\%  & 87\%  & 96\% \\
	% 			\bottomrule
	% 		\end{tabular}
	% 		\label{tab:edge}
	% 		\vskip -1em
	% 	\end{table} 
	
	\smallskip\noindent\textbf{Main Results.} 
	Table~\ref{tab:overall} shows the node classification performance of all comparing methods on different datasets.
	Specifically, the first row of Table~\ref{tab:overall} presents the positive rate of original graphs and LA-graphs, while
	the other rows present the performance of the baseline GCN models and their LA enhanced versions. 
	The results verify that our proposed LAGCN framework can improve the node classification performance of existing GCN models considerably. In other words, increasing the positive ratio of the underlying graph can lead to better classification performance for GCN models, i.e., Theorem~\ref{the:origin}.
	Moreover, the results also show that 
	1) the original GAT outperforms the original GCN and the original SGC; however, both the LA-SGC and the LA-GCN perform better than the original GAT, which indicates that the LAGCN framework is more effective than the attention mechanism;
	2) the LA-GAT outperforms the original GAT, which indicates that LAGCN can complement the attention mechanism to reach a better performance. 
	%It is also noteworthy that the LA-GCN, the LA-SGC, and the LA-GAT outperforms their original versions by almost $10$ percent on the Pubmed dataset.
%	3) for the sampling based methods, their LA enhanced versions perform better than the original ones.
	%; meanwhile, the LA-ASGCN outperforms all the compared methods remarkably.
	
%	\begin{figure*}[t]
%		%\vskip -0.5em
%		\begin{center}
%			\centerline{\includegraphics[width=1.12\linewidth,trim=2cm 0 0 0,clip]{ablation}}
%			\vskip -0.5em
%			\caption{The performance of the GCN models under different graphs. Filter-graphs are generated by only using the filtering process; add-graphs are generated by only using the adding process; LA-graphs are generated by using both of them.} 
%			\label{fig:addfilt}
%		\end{center}
%		\vskip -2em
%	\end{figure*}
%	\begin{figure*}[htbp]
%		% 		\vspace{-1em}
%		\begin{center}
%			\centerline{\includegraphics[width=1.03\linewidth,trim=0 0 0 0,clip]{gr_predict}}
%			
%			\caption{The performance of LA-SGC when varying the values of $p-q$ and $p_{pre}$. The results on the left and the middle panel are generated with the filter-graph and the add-graph, respectively. The results on the right panel are generated with the LA-graph.}
%			\label{fig:gr_predict}
%		\end{center}
%		\vskip -1em
%	\end{figure*}
%	\vspace{0.2em}
%	\subsubsection{Ablation Study}
%	\label{sec:ablation}
    
\begin{figure}[t]
	\vspace{-1em} 
	\begin{center}
		\centerline{\includegraphics[width=\linewidth,trim=0 0 0 0,clip]{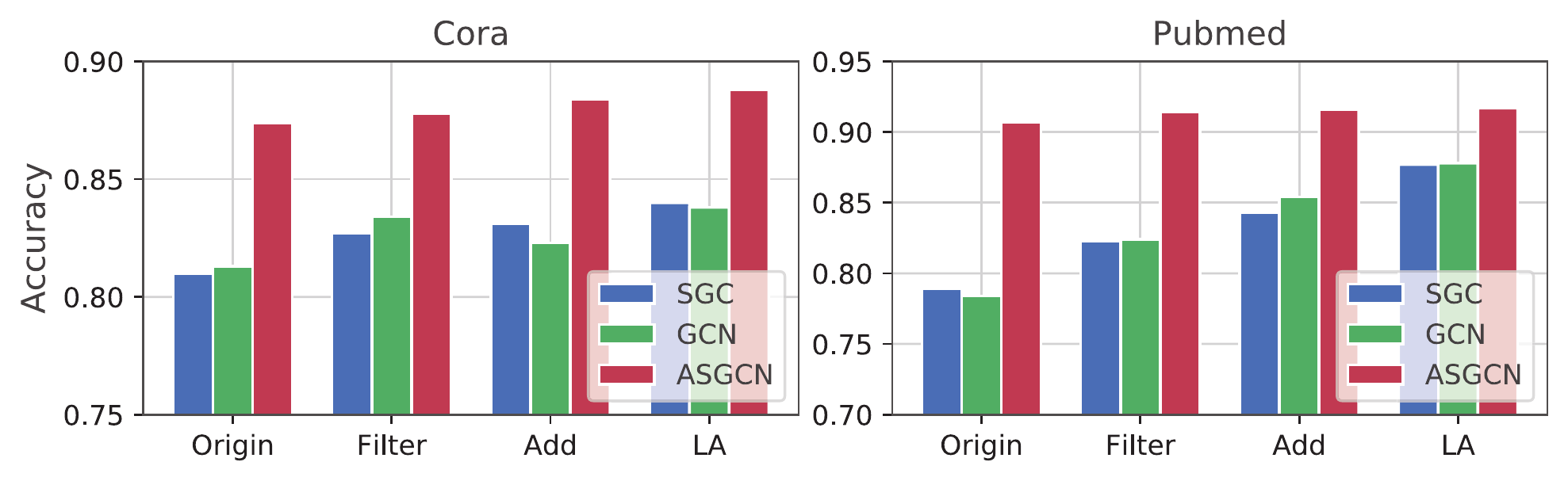}}
		\vspace{-1.5em}
		\caption{Influence from the adding and filtering process.} 
		\label{fig:addfilt}
	\end{center}
	\vskip -2em
\end{figure}

	\smallskip\noindent\textbf{Influence from the adding and filtering process.} In Figure~\ref{fig:addfilt}, we evaluate the influence of the adding and filtering process, respectively. The filter-graph only filters the negative edges; the add-graph only adds more positive edges; the LA-graph both filters the negative edges and adds more positive edges.
	This figure demonstrates that both the filtering and the adding process can enhance the performance of the GCN models. Meanwhile, they can complement each other to reach the best performance. Note that similar conclusions can also be drawn from the other datasets, we omit the results here due to space limit.
	
\begin{figure}[t]
	\vspace{-0.7em} 
	\begin{center}
		\centerline{\includegraphics[width=\linewidth,trim=0 0 0 0,clip]{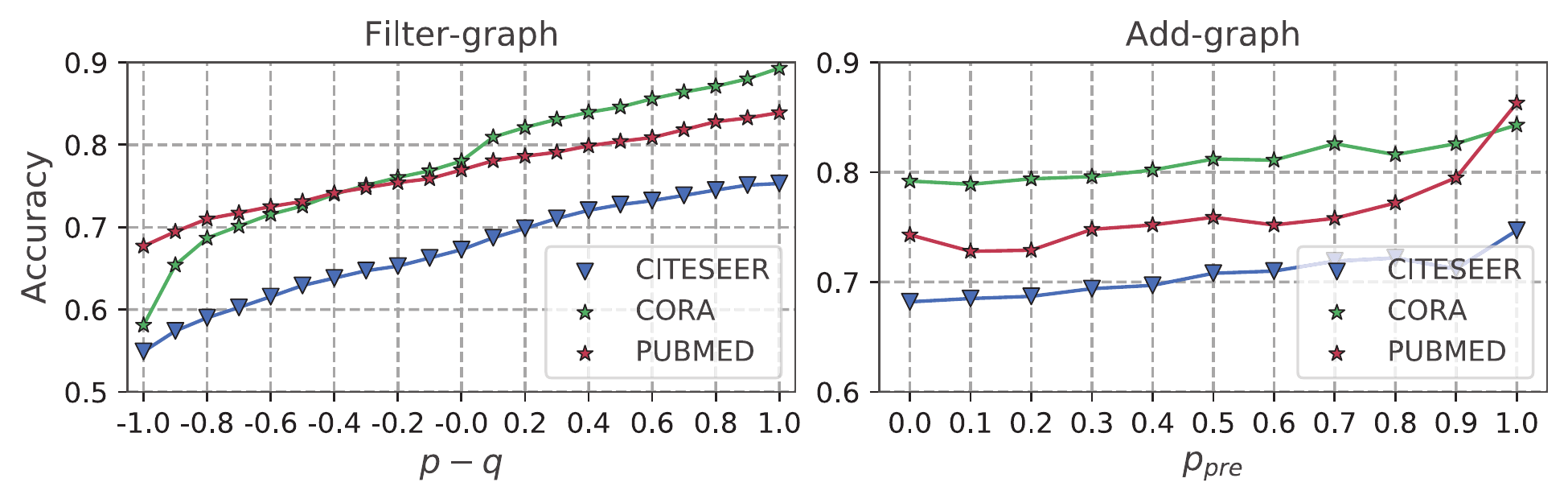}}
		\vspace{-1.5em}
		\caption{Influence from the edge classifier.} 
		\vspace{-2em}
		\label{fig:gr_predict}
	\end{center}
\vspace{-0.5em}
\end{figure}

	\smallskip\noindent\textbf{Influence from the edge classifier.} In Figure \ref{fig:gr_predict}, we study the influence of the edge classifier by artificially modifying the $p-q$ and $p_{pre}$ with the ground truth testing labels and evaluating it with LA-SGC.
	The sub-figure on the left panel shows that the performance of the LA-models is positively correlated to the value of $p-q$, which verifies Theorem~\ref{the:filter}.
	The sub-graph on the right panel shows that the performance of LA-models is positively correlated to the value of $p_{pre}$, which verifies Theorem~\ref{the:add}.
%	The sub-graph on the right panel shows how the performance of LA-SGC on Cora varies while modifying the values of $p-q$ and $p_{pre}$. The result shows that the lines corresponding to different $p_{pre}$ values are almost in parallel, which indicates that the adding process and the filtering process have little influence on each other.

	\smallskip\noindent\textbf{Performance under low positive ratio.}
	We add 5 different-label neighbors to each node in Cora, Citeseer and 15 different-label neighbors in Pubmed so as to artificially decrease their positive ratio from $85\%$, $74\%$, $80\%$ to $ 30\%$, $24\%$, $34\%$, respectively, and test the performance of GCN and GAT.
	The LA classifier used in LA-GCN and LA-GAT are trained with the raw features and can improve the positive ratio back to $ 67\%$, $74\%$, $95\% $, respectively. 
%	and test the performance of LA-GCN and LA-GAT.
	The results in Table~\ref{tab:low-homophily-ratio} show that LA-GCN and LA-GAT are much more robust than GCN and GAT when the graph has a lower positive ratio.
%	GCN and GAT generate a worse performance  The superior performance of  proves that LAGCN helps even the graph has a low positive ratio.
	%This may due to that the convolution and attention mechanism have limited effect when the nodes in a graph have few positive neighbors. 
	% 	On the other hand, 
	%The superior performance of LA-GCN and LA-GAT proves that the LAGCN can help existing GCNs to achieve much better performance by refining the graph topology, especially when the graph has a low positive ratio.
	% 	\vspace{-1em}
	
% \vspace{-0.5em}
\section{Conclusion}
% \vspace{-0.3em}
	In this paper, we propose the LAGCN framework, which can increase the positive ratio of the learning graph by training an edge classifier to filter the negative neighbors and add new positive neighbors for each node in the graph. 
	Experimental results verify that existing GCN models can directly benefit from LAGCN to improve their node classification performances.
% \vspace{-0.5em}
\section{Acknowledgement}
% \vspace{-0.3em}
This work is supported in part by the National Natural Science Foundation of China~(Grand Nos. U1636211, 61672081, 61370126) and 2020 Tencent Wechat Rhino-Bird Focused Research Program and China Postdoctoral Science Foundation~(No. 2020M670337).
%	Future works can follow LAGCN to study how to adapt LAGCN to other type of datasets such as heterogenious graph or multi-label datasets.
	
\vspace{-0.7em}
	% 	\newpage
	% 	\bibliographystyle{named}
	\bibliographystyle{plain}
	\bibliography{references}

\begin{thebibliography}{10}

\bibitem{bhagat2011node}
Smriti Bhagat, Graham Cormode, and S~Muthukrishnan.
\newblock Node classification in social networks.
\newblock In {\em Social network data analytics}, pages 115--148. Springer,
  2011.

\bibitem{chen2018fastgcn}
Jie Chen, Tengfei Ma, and Cao Xiao.
\newblock Fast{GCN}: fast learning with graph convolutional networks via
  importance sampling.
\newblock {\em ICLR}, 2018.

\bibitem{grover2016node2vec}
Aditya Grover and Jure Leskovec.
\newblock node2vec: Scalable feature learning for networks.
\newblock In {\em SIGKDD}, pages 855--864. ACM, 2016.

\bibitem{hamilton2017graphsage}
Will Hamilton, Zhitao Ying, and Jure Leskovec.
\newblock Inductive representation learning on large graphs.
\newblock In {\em NIPS}, 2017.

\bibitem{huang2018asgcn}
Wenbing Huang, Tong Zhang, Yu~Rong, and Junzhou Huang.
\newblock Adaptive sampling towards fast graph representation learning.
\newblock In {\em NIPS}, 2018.

\bibitem{kipf2016gcn}
Thomas~N Kipf and Max Welling.
\newblock Semi-supervised classification with graph convolutional networks.
\newblock {\em arXiv preprint arXiv:1609.02907}, 2016.

\bibitem{li2018laplacian}
Qimai Li, Zhichao Han, and Xiao-Ming Wu.
\newblock Deeper insights into graph convolutional networks for semi-supervised
  learning.
\newblock In {\em AAAI}, 2018.

\bibitem{perozzi2014deepwalk}
Bryan Perozzi, Rami Al-Rfou, and Steven Skiena.
\newblock Deepwalk: Online learning of social representations.
\newblock In {\em SIGKDD}, pages 701--710. ACM, 2014.

\bibitem{rong2019dropedge}
Yu~Rong, Wenbing Huang, Tingyang Xu, and Junzhou Huang.
\newblock Dropedge: Towards the very deep graph convolutional networks for node
  classification.
\newblock {\em ICLR}, 2020.

\bibitem{velivckovic2017gat}
Petar Veli{\v{c}}kovi{\'c}, Guillem Cucurull, Arantxa Casanova, Adriana Romero,
  Pietro Lio, and Yoshua Bengio.
\newblock Graph attention networks.
\newblock {\em ICLR}, 2019.

\bibitem{wang2014mmrate}
Senzhang Wang, Xia Hu, Philip~S Yu, and Zhoujun Li.
\newblock Mmrate: inferring multi-aspect diffusion networks with multi-pattern
  cascades.
\newblock In {\em KDD}, 2014.

\bibitem{wu2019sgc}
Felix Wu, Tianyi Zhang, Amauri Holanda~de Souza~Jr, Christopher Fifty, Tao Yu,
  and Kilian~Q Weinberger.
\newblock Simplifying graph convolutional networks.
\newblock {\em ICML}, 2019.

\bibitem{zeng2019graphsaint}
Hanqing Zeng, Hongkuan Zhou, Ajitesh Srivastava, Rajgopal Kannan, and Viktor
  Prasanna.
\newblock Graphsaint: Graph sampling based inductive learning method.
\newblock {\em ICLR}, 2020.

\bibitem{zhang2018gaan}
Jiani Zhang, Xingjian Shi, Junyuan Xie, Hao Ma, Irwin King, and Dit-Yan Yeung.
\newblock Gaan: Gated attention networks for learning on large and
  spatiotemporal graphs.
\newblock {\em WWW}, 2018.

\end{thebibliography}
\newpage

\end{document}